# Driver-Specific Risk Recognition in Interactive Driving Scenarios using Graph Representation

# Driver-Specific Risk Recognition in Interactive Driving Scenarios using Graph Representation

Jinghang Li, Chao Lu, Penghui Li, Zheyu Zhang, Cheng Gong, Jianwei Gong, *Member, IEEE*

*Abstract*—This paper presents a driver-specific risk recognition framework for autonomous vehicles that can extract inter-vehicle interactions. This extraction is carried out for urban driving scenarios in a driver-cognitive manner to improve the recognition accuracy of risky scenes. First, clustering analysis is applied to the operation data of drivers for learning the subjective assessment of risky scenes of different drivers and generating the corresponding risk label for each scene. Second, the graph representation model (GRM) is adopted to unify and construct the features of dynamic vehicles, inter-vehicle interactions and static traffic markings in real driving scenes into graphs. The driver-specific risk label provides ground truth to capture the risk evaluation criteria of different drivers. In addition, the graph model represents multiple features of the driving scenes. Therefore, the proposed framework can learn the risk-evaluating pattern of driving scenes of different drivers and establish driver-specific risk identifiers. Last, the performance of the proposed framework is evaluated via experiments conducted using real-world urban driving datasets collected by multiple drivers. The results show that the risks and their levels in real driving environments can be accurately recognized by the proposed framework.

## I. Introduction

According to the survey carried out by National Highway Traffic Safety Administration (NHTSA), about 94% of traffic accidents are caused by human drivers. In these accidents, unsafe driving behavior is responsible for 95% of fatal traffic accidents [1, 2]. It is expected that the development of Advanced Driver Assistance Systems (ADASs) will prevent such accidents from happening and thus save numerous lives. To achieve this goal, the ADASs should be able to accurately recognize the potential risks in driving scenes and timely warn or correct the drivers of their unsafe driving behavior.

However, risk recognition still remains a bottleneck problem in the development of ADASs. On the one hand, it is difficult to understand and model the dynamic traffic scenes, where different traffic participants, road conditions and traffic rules jointly affect the risks involved in the driving scenes. On the other hand, different drivers tend to have different driver-specific standards for recognizing driving risks because of their differences in driving experiences and personalities. This kind of recognition differences will result in human-machine conflicts that might perturb the drivers and lead to severe accidents.

In the past few decades, many studies on various risk assessment indicators have been carried out. These indicators include the widely applied TTX family (time-to-x) [3], THW (time headway) [4], and CCP (current car position) [5]. These straightforward methods usually assess the driving risks directly using the geometric or kinematic relations between the ego and surrounding vehicles. They are computationally efficient and have been applied in many scenarios. However, their oversimplified scene modeling fails to capture complex relationship between multiple dynamic traffic participants. Therefore, use of these indicators is limited to only a few specific scenarios with a limited generalization ability.

In recent years, learning-based methods have demonstrated their significant advantages in scene understanding and pattern recognition. These methods can be used to properly model the complex traffic scenes. Many related techniques have been proposed for motion prediction [6, 7], abnormal driving detection [8], risk assessment [9, 10], and other tasks [11-13]. One example was shown in [12], where statistical learning was used to symbolize driving data and extract key points from the dataset to identify lane-change risks. Another learning-based method was proposed in [10] for collision risk evaluation. In these references, an intention identification model was set up via long short-term memory (LSTM) networks to identify the intention possibility of the surrounding vehicles. A driving safety field was calculated to output the potential risk. Although these learning-based methods were effective in representing the complex traffic scenes, they failed to capture the interactive relationships between different participants. Indeed, the interactive influence of traffic participants is one of the most important factors affecting the risk levels of a driving scene.

As compared to the traditional data representation methods, the graph representation model (GRM) is regarded as an ideal technique to capture the interactive relationships between different objects. Thanks to this advantage, the graph models have increasingly been gaining attention in the field of human motion recognition [14], image scene understanding [15], recommender systems [16], and social network analysis [17].

This work was supported by the National Natural Science Foundation of China under Grant 61703041.

J. Li, C Lu, Z. Zhang, C. Gong, J. Gong are with the School of Mechanical Engineering, Beijing Institute of Technology, Beijing 100081, China (e-mail: 3120180346@bit.edu.cn; chaolu@bit.edu.cn; 3120190429@bit.edu.cn; chenggong@bit.edu.cn; gongjianwei@bit.edu.cn).

Penghui Li is with Key Laboratory of Transport Industry of Big Data Application Technologies for Comprehensive Transport, the School of Traffic and Transportation, Beijing Jiaotong University, China (e-mail: penghui@bjtu.edu.cn).

(Corresponding authors: Chao Lu, Penghui Li)

In the field of intelligent transportation, the vehicles, pedestrians, and other traffic participants on the road can be seen from a bird's eye view as a set of nodes on a graph and the spatial geometric relations between them can be considered as edges. Therefore, compared with traditional data representation method in a vector form, the data structure of the graph can simultaneously express the position relationship between the dynamic and static elements in the traffic environment, as well as their implicit interaction in the form of edges. This superior relation representation has been proven effective in road agents' behavior recognition [18], and trajectory prediction [19]. However, its application on risk assessment has not been widely explored.

Apart from the problem in interactive relationship modeling, unified understanding of risk levels between the ADASs and drivers is the prerequisite of harmonic human-machine interaction. However, due to the differences of driver experiences and personalities, a perfectly trained risk recognition model of ADASs for specific drivers will cause human-machine conflicts when applied to other drivers because of different understandings of the risk. These conflicts may lead to faulty recognition or cause the driver to ignore warnings [20-22]. A unified and objective risk level assessment standard should be implemented to solve this problem. A human-generated risk level label was proposed to reflect the drivers' assessment of a risky scene.

In [23], the risk level labels were obtained by showing study participants driving videos and collecting their assessment of the scene's risk level. Furthermore, existing research on driver risk perception includes questionnaire surveys or self-reported information from drivers. The driving data in high-risk situations were usually collected from a driving simulator [23-26]. Among these methods, risk perception is the output of human cognition of the environment. However, in order to provide assessment of the risk level of the scene, the driver should have a real sense of environmental information such as the visual information, vehicle acceleration, and the relative motion between the host vehicle and the surrounding vehicle. Instead of subjective judgement, the driver's assessment of the risk level of the scene can also be reflected in the actual operation data, i.e., the change of the vehicle deceleration. In [27], deceleration was proposed as an index to quantify the risky scenarios.

In this paper, a framework for driver-specific risky scene recognition based on the GRM is proposed. In order to evaluate the effectiveness of the proposed framework, the training and validation datasets are collected from real driving environments. The environments include crowded urban scenes and detailed CAN bus information. The datasets are collected with a high data sampling frequency to meet the needs of data-driven human-like driving research. In addition, the node and edge data structure of the graph model is utilized to simultaneously recognize risky scenes and introduce the interaction behavior between vehicles. The goal of this paper is to propose a new framework for risky scenes recognition based on graph representation learning and investigate the important role of vehicle interaction.

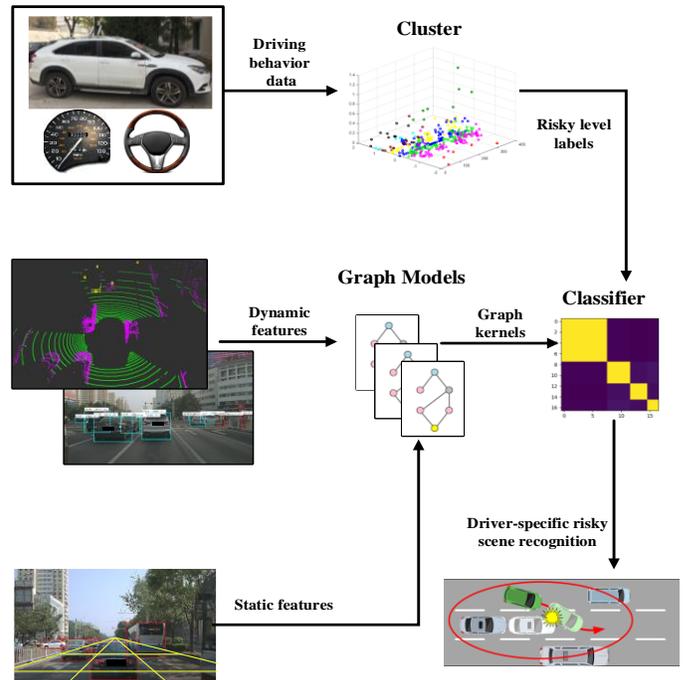

Fig. 1. The proposed framework.

The rest of this paper is organized as follows. The proposed framework for risky scene recognition is presented in Section II. Section III describes the risky scene recognition problem and provides a detailed definition of the proposed method. A group of experiments using realistic driving data are designed and detailed in Section IV to validate the proposed method. Finally, conclusions and future work are presented in Section V.

## II. PROPOSED FRAMEWORK

In this paper, a learning-based framework for risky scenes recognition is proposed. Following are its main features:

1) The framework can learn the subjective tendencies of risk perception from different drivers.

2) The framework can model the complex traffic environment of real driving scenes and represent the dynamic relationships between different traffic participants.

3) The framework can recognize the risk present in driving scenes in a driver-specific manner and seek to minimize the human-machine conflict.

Fig. 1 illustrates the structure of this framework. First, the driver operation data are analyzed by clustering to explore the subjective risk-cognitive pattern of different drivers. The risk level labels are obtained through clustering for the corresponding scenes. Second, real driving scenes are extracted and modeled through the GRM. The extracted model contains key information such as the dynamic and static feature information of the scenes and the interaction between the traffic participants. Last, the extracted graph model is combined with clustered risk level labels to train the risk recognition model. The trained model corresponds to the subjective risk-cognitive pattern of the driver. The model can simulate the risk evaluation process of drivers by using the key information of the current driving scene and providing the corresponding personalized risk level of the scene.

A common and convenient method for obtaining the subjective risk-cognitive patterns of different drivers is to either send questionnaires to volunteers or organize them to subjectively score the risk level of collected video clips of real driving scenes. However, these methods can be labor-intensive and time-consuming in practice. In fact, the driving response, such as slowing down or braking sharply, implicitly shows the judgment of the driver regarding the risk level of a particular scene. Therefore, a more accurate and personalized subjective risk perception label can be obtained by analyzing the operational data based on the actions of driver. The clustering algorithm is a typical data analysis method that can handle large-scale data and place data with similar statistical characteristics in the same group. The operational features that are the most sensitive to the risk evaluation of drivers are determined in order to utilize the clustering algorithm for analyzing driver operation data. This is further described in detail in Section III. The selected feature data are then clustered and analyzed to obtain risk level labels of different drivers for driving scenes.

It is necessary to include all key information affecting driving behavior in the driving scenes modeling process. This results in effective training of the recognition model for mapping the relationship between the driving scenes and the risk level labels of drivers. While driving, the driver observes the real-time information of dynamic and static features in the environment, such as lanes and locations of other traffic participants. However, he/she should also consider the interaction between the host vehicle and the surrounding traffic participants to recognize the risk in the current scene. The GRM has been shown to be very effective in modeling the interactions between different entities. In graph representation, the dynamic and static elements are modeled as graph nodes, and the node labels, respectively. Furthermore, the interactions between different factors are modeled as the node connection edges, which fully represents multiple and changing number of dynamic and static features as well as their interaction relations. Modeling of driving scenes using the GRM not only provides an effective representation of the changing number of traffic participants during the driving process but also represents features and interactions in a unified model. Therefore, the GRM-based model can represent an enhanced understanding of the driving scene.

As the driving scenes and their corresponding risk level labels are known, the subjective risk assessment pattern of the driver can be learned through machine learning. Support vector machine (SVM) is a commonly used machine learning algorithm, which can find the decision boundary between the sample data and learn the correspondence between the sample features and labels. Therefore, the SVM can be applied to learn the mapping relationship between different scenes and the corresponding subjective risk level labels of driver. In addition, it can obtain a risky scene recognition model that fits the subjective risk assessment pattern of the driver.

## III. RISK LABEL GENERATION

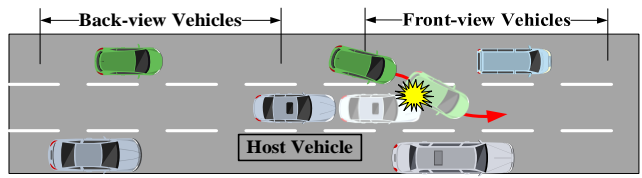

Fig. 2. The lane change of front vehicle in heavy urban traffic scene.

The personalized recognition of risky driving scenes focuses on learning the pattern of human drivers' understanding of the risky scenes. Furthermore, this pattern is used to train a risk level classifier based on the input of driving scene features. The understanding of risk by the driver can be reflected in his/her driving operation, as the driver avoids the risk by maneuvering the vehicle. Therefore, the operation data should be analyzed to learn the risk understanding patterns of drivers and provide risk labels for classifier training.

In this paper, a risky scene is defined as a potential collision between the host vehicle and surrounding vehicles, as shown in Fig. 2. The drivers usually avoid risks through quick evasive maneuvers, i.e., emergency braking and/or steering. Therefore, many studies in the field of driving behavior research, such as that performed by Guo et al. [27, 28], have used the evasive maneuvers recognition method to calculate the likelihood of near collisions between vehicles. However, the method is not commonly applied for risky scene recognition.

It can be found by reviewing previous accident data that the drivers prefer to use quick braking rather than emergency steering to avoid risks [29]. This choice can be understood by considering that the emergency steering may lead to a collision between the host vehicle and the oncoming vehicle from the rear. Therefore, the longitudinal acceleration (AccX) of host vehicle is first selected as **Feature One**. However, the lateral and longitudinal operations of a driver modify the vehicle state together, and the high-dimensional data also contain a higher amount of implicit information [30, 31]. Thus, the high-dimensional features containing all driver operation data are selected as **Feature Two** for comparison.

Clustering is the most common unsupervised learning method for exploring data structures and statistical data analysis. In this paper, the classical K-means clustering algorithm (KMC) is initially chosen. Subsequently, considering the high-dimensionality of **Feature Two**, the spectral clustering algorithm (KPCA-KMC), which has a stronger clustering ability for high-dimensional data, is selected. The KPCA-KMC maps high-dimensional data to low-dimensional space and reduces the influence of irrelevant dimensions on the clustering performance, resulting in accurate clustering of high-dimensional data.

### A. K-Means clustering (KMC)

The K-means algorithm is a popular method for cluster analysis in the field of data mining [32]. Its purpose is to partition *n* points, which can be one observation or one instance of a sample, into *k* clusters, where each point belonging to the cluster corresponds to the nearest mean with respect to the cluster center. These *k* clusters correspond to *k* clustering labels. The objective function of KMC is expressed as follows:

$$dist_{ed}(a_{h,i}, a_{h,j}) = \|a_{h,i} - a_{h,j}\|_2 = \sqrt{\sum_{i,j=1}^{n}|a_{h,i} - a_{h,j}|^2} \ (i,j \in n) \ (1)$$

where $a_{h,i}$ and $a_{h,j}$ are the driving behavior data of the host vehicle at time $i$ and $j$, respectively, with $i, j = 1, 2, \ldots, n$ representing the time series length.

### B. Spectral clustering (KPCA-KMC):

The KPCA is first utilized to achieve dimensionality reduction for high-dimensional input data, and help determine the number of clusters, $K$, by analyzing the principal components [33]. In the KPCA, the input $\mathbf{S}$ is first projected onto a high-dimensional feature space $\phi$, which enables the original data to become linearly separable. Subsequently, the conventional PCA is applied to $\phi(\mathbf{S})$, resulting in a low-dimensional data projection $\hat{\phi}(\mathbf{S})$ on the principal components space. This projection accounts for most of the variance in the original signal. After dimensionality reduction using the KPCA, the KMC is applied to partition the observations $\hat{\phi}(\mathbf{S})$ into $K$ clusters $\mathbf{C} = \{c_1, c_2, \ldots c_k\}$ by minimizing the intra-cluster sum of squares.

We have described two common clustering algorithms in the previous sub-sections. During the process of using the clustering algorithms, the number of classes, i.e., K-values for data clustering is selected manually. Therefore, evaluation methods are needed to judge the clustering results, which are described next.

### C. Residual Sum of Squares

In this paper, the residual sum of squares (RSS) combined with the elbow principle is used to estimate the performance of clusters for selecting the best number of clusters $K$. The RSS represents the sum of squares of the errors within each cluster. For a data point $\mathbf{s}_i$ in a dataset $\mathbf{D} = \{x^i \in R\}_{i=1}^{N}$, the RSS can be calculated as follows:

$$\text{RSS} = \sum_{k=1}^{K} \sum_{x^i \in C_k} \|\mathbf{s}_i - \mathbf{\mu}_k\|^2 \quad (2)$$

The elbow principle is used to determine the number of categories for clustering the data set, i.e., the value of $K$. As $K$ increases, the dataset is divided more finely and the degree of clustering in each cluster gradually increases, leading to a smaller RSS. When $K$ is less than the optimal value, the decrease in RSS will be significant when $K$ increases. When $K$ is assigned the optimal value, the decline in RSS becomes smaller. Finally, it tends to level off as $K$ increases. Hence, the value of $K$ corresponding to the elbow is the optimal number of clusters.

### D. Silhouette Coefficient

The performance of the clustering results is evaluated by using the silhouette coefficient (SC) in this paper for selecting the features that best represent the risk perception of drivers. The SC combines the cohesion and separation of clustering to show how similar a data point is to its own cluster compared to other clusters. For the data point $\mathbf{s}_i$ belonging to the cluster $C_i$, its silhouette value $Sil(\mathbf{s}_i)$ can be calculated via the following equations (3), (4), (5):

$$a(x^i) = \frac{1}{|C_i| - 1} \sum_{x^j \in C_i, i \neq j} d(x^i, x^j) \quad (3)$$

$$b(x^i) = \min_{k \neq i} \frac{1}{|C_k|} \sum_{x^j \in C_k} d(x^i, x^j) \quad (4)$$

$$Sil(\mathbf{s}_i) = \frac{b(\mathbf{s}_i) - a(\mathbf{s}_i)}{\max\{a(\mathbf{s}_i), b(\mathbf{s}_i)\}} \quad (5)$$

where $a(\mathbf{s}_i)$ denotes the average distance between the data point $\mathbf{s}_i$ and all other data points in the same cluster, $b(\mathbf{s}_i)$ denotes the minimum average distance between a data point and all points in any other cluster, and $d(\mathbf{s}_i, \mathbf{s}_j)$ is the Euclidean distance between data point $\mathbf{s}_i$ and data point $\mathbf{s}_j$.

The above definitions show that the silhouette values should range between -1 to +1. A higher silhouette value indicates better matching of the data points to their own clusters. A silhouette value of 1 for the data point $\mathbf{s}_i$ indicates that it is very compact within the cluster it belongs to and is far from other clusters. A value of -1 indicates the worst clustering performance. The SC is the mean of the silhouette values of all samples and is a reliable measure to determine whether the current clustering result is valid and reasonable.

## IV. GRAPH REPRESENTATION MODEL

A scene model containing critical information that influences the driving risk evaluation is needed to train an effective risk level classifier. Traditional vector representation modeling (VRM) approaches detect the risk situation by separately focusing on the dynamic properties of the surrounding vehicles. However, in the crowded urban environment, the risk level of the scene can change rapidly as multiple traffic participants continuously influence one another under the constraint of road infrastructures. Although the traditional vector representation includes the dynamic information of different traffic participants, it neglects the influence of the interaction between the traffic participants and the constraint of road infrastructures. Besides, as the number of traffic participants changes during the driving process, the traditional vector representation methods cannot effectively adapt to the changing number of traffic participants as the dimension of the feature space is usually fixed.

To overcome these obstacles and attain better understanding of the driving scene, the GRM is chosen due to its capability of modeling interconnecting relationship among multiple objects, and adaptivity to the changing number of objects. Due to these features, the GRM is widely used in fields such as chemistry and biology, where it uses nodes and edges to represent the atoms and the relationships among them. As presented in Fig. 3, a labeled undirected graph is defined as $G = (V, E)$ consisting of a set of vertices $V$ and a set of edges $E \in V \times V$ that connect different pairs of vertices. The graph is endowed with a function $\ell : V \cup E \to L$ that assigns labels to the vertices of the graph from a discrete set of labels $L$.

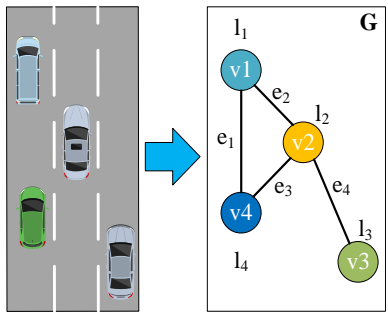

Fig. 3. Traffic scene representation by a graph.

Modeling of driving scenes using the GRM requires finding the correspondence between the features in the scenes and the definitions in the graph model. According to [34], the driving scene includes dynamic and static features and the interactions between them. The dynamic features are mainly vehicles, while the static features contain geometric information of urban roads and traffic signage information, such as lane indicators. Therefore, the modeling process contains 1) scene static feature extraction, 2) scene dynamic feature extraction, and 3) definition of interaction between the vehicles.

To obtain an effective representation of driving scenes that contains critical risk-related information, the key elements of a driving scene should first be defined. An effective graph model is expected to contain all critical features and represent them properly. Therefore, embedding the features into the graph model shall also be discussed in the following section. Specifically, the extraction and graph modeling of three features, namely dynamic features, static features, and the interactions between them, are demonstrated. Additionally, two graph kernel methods are introduced to measure the similarity between the graph models. The widely used SVM classifier is trained for risky scene recognition.

*A. Scene static feature extraction*

Unlike off-road environment, in structured urban roads, the road surface and lane indicators define the area in which the vehicles can travel. At the same time, traffic rules stipulate that the motor vehicles can only drive in lanes except when changing lanes. Due to this, the trajectories of vehicles on the road are restricted to static features of the road and are highly structured. Therefore, in order to consider this effect of static feature, the lane line information is introduced in the graph model.

Based on the above discussion, vehicles can be naturally defined as nodes of a graph model from the aerial view of urban roads. Thus, the collected urban driving dataset should be transformed into a bird's eye view based on the host vehicle, i.e., the data collection platform. The host vehicle is also modeled as a node. In this case, the interaction between the host vehicle and the surrounding vehicles is also included in the graph. As the aerial view is generated with the data collection platform as the coordinate origin, the node of host vehicle is always centered below the aerial view as shown in Fig. 4. The x-direction of road surface is divided into 3 parts based on the number of lanes on road. Specifically, the lane indicators detected by computer vision are used to divide the road surface of each aerial view into three lanes along the x-direction. In the y-direction, the farthest detection distance of the on-board sensor is 100 m. A car length of about 5 m on the city road and the following distance of approximately 5 m are considered. Based on these dimensions, the y-direction is divided into 10 regions with a 10 m interval, and the road surface is finally divided into a 3x10 grid. Each grid is coded sequentially starting from one.

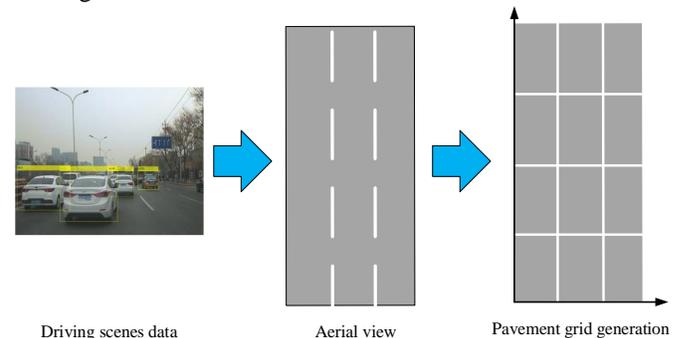

Driving scenes data    Aerial view    Pavement grid generation

Fig. 4. Graph model node definition.

*B. Scene dynamic feature extraction*

As discussed earlier, the dynamic scene features mainly focus on the relative positions of the host vehicle and the surrounding vehicles. By performing multi-sensor information fusion, the positions of the surrounding vehicles can be collected in the coordinate system of the host vehicle. Based on the grid structure of the graph, the continuous position of the vehicle is discretized as the occupancy of grid. Specifically, the vehicles within the detected field are coded as nodes. The nodes in each aerial view are labeled according to the grid where they are located, and the node labels of the graph model are obtained as shown in Fig. 5.

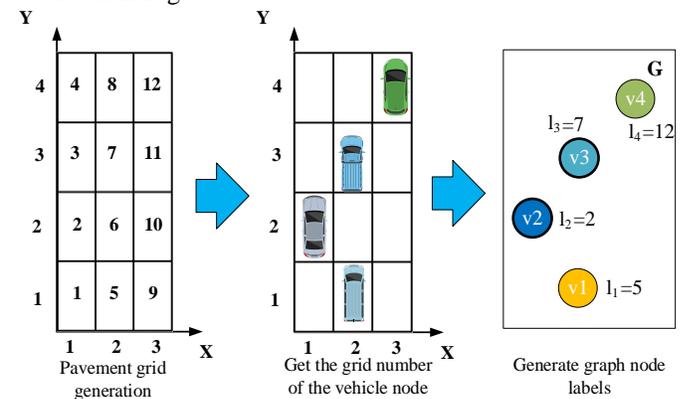

Pavement grid generation    Get the grid number of the vehicle node    Generate graph node labels

Fig. 5. Graph node label definition.

*C. Definition of interaction between vehicles*

In the graph model, two nodes can be connected by an edge, which means that the two nodes are related. Examples of the nodes and edges include the connecting bond of two elements in a molecular structure, or the social relationship of two people in a social network. In this paper, the edges in the graph model represent the potential collision relationship between the host vehicle and the surrounding vehicles. Since the smaller distance between the vehicles results in higher probability of vehicle collision, the distance is considered as an important metric to approximate interactions. Therefore, a vehicle that exists within a 3x3 grid centered on vehicle A is considered to have an edge connection with vehicle A. In addition, since the collision is equal for both vehicles, the edges defined in this paper are two-way and labeled as 1. Fig. 6 shows the defined model.

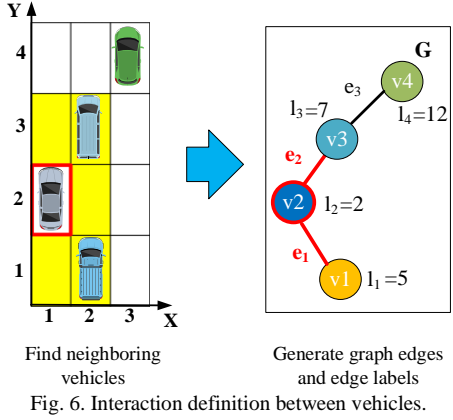

Fig. 6. Interaction definition between vehicles.

*D. Classifier Training*

The similarity between the graph should be calculated prior to training the risky scenes recognition classifier. The graph is complex nonlinear structural data, which cannot be classified using the simple linear SVM model. In this paper, the graph kernel method is used to project the inner product between graphs into a high-dimensional Hilbert space, and subsequently a linear SVM model is used for training. Two graph kernel algorithms based on different computational methods are compared: the shortest path graph kernel (SPGK) based on path calculation and the neighborhood hash graph kernel (NHGK) based on neighborhood aggregation calculation.

- **Shortest Path Graph Kernel (SPGK)**

The SPKG decomposes each graph in a graph structure data into a combination of shortest paths [35]. It compares different graphs based on the length of the shortest path and the labels of the path endpoints. The specific calculation steps are: First, the input graph is transformed into a shortest path graph $S = (V, E_S)$, i.e., given an input graph $G = (V, E)$, where $V$ and $E$ are the sets of nodes and edges of $G$. The SPGK is used to create a shortest path graph based on the input graph $G$, where the set of nodes $V$ contained in $S$ is the same as $G$. The set of edges $E_S$ of $S$ represents the shortest path between all nodes in $G$ and is a subset of $G$. The mathematical definition of the algorithm is as follows:

$$k(G, G') = \sum_{e \in E} \sum_{e' \in E'} k_{walk}^{(1)}(e, e') \tag{6}$$

where $k_{walk}^{(1)}(e, e')$ is a semi-positive definite kernel on the side of the walk path of length 1.

- **Neighborhood Hash Graph Kernel (NHGK)**

The NHGK is a typical representative of the neighborhood aggregation algorithm, also known as the message passing algorithm [36]. The algorithm measures the similarity between different graphs by updating the node labels of the input graph and counting their number of common labels. The kernel replaces discrete node labels with fixed-length binary arrays. Subsequently, a logical operation is used to update the node labels so that they contain information about the adjacency structure of each node. Using the neighborhood hashing operation $(1, 2, ..., h)$ times on the nodes of the two input graphs $G$ and $G'$, the updated graphs based on the two input graphs are $G_1, G_2, ..., G_h$ and $G_1', G_2', ..., G_h'$. The distance metric of the two input graphs is calculated as follows:

$$k(G, G') = \frac{1}{h} \sum_{i=1}^{h} \frac{c}{|V| + |V'| - c} \tag{7}$$

where c is the number of labels that the two graphs have in common. This distance metric is commonly used as a similarity measure between sets of discrete values and has been shown to be positive semidefinite [37].

Based on the work presented in [38], the LIB-SVM implementation is used for solving the risky scenes recognition problem. The LIB-SVM is an easy to use, fast and effective SVM pattern recognition and regression package. It not only provides a compiled executable file to run on Windows systems, but also provides the source code that can be easily improved, modified, and applied on other operating systems.

## V. EXPERIMENTS

In this section, we evaluate our framework in four steps. First, the data acquisition platform is used to collect operational data of three drivers and the corresponding driving scene features in the real world. We model the dynamic and static features of the driving scene and the interaction between the vehicles using the GRM, as described in Section IV. Second, the clustering analysis is conducted based on the Operational data of drivers for generating the driver-specific risky labels. Two clustering methods are compared using the collected operational data of drivers to select the features that best respond to the risk perception of the driver. Third, after generating the driver-specific risky labels, GRM-based and VRM-based risky scenes recognitions are performed for comparison. In this work, the GRM-based experiments and VRM-based experiments are defined as follows.

1. GRM-based experiments: The driving scene features are represented by the GRM during the training of SVM model.
2. VRM-based experiments: The driving scene features are represented by the traditional VRM during the training of SVM model.

The third step is mainly carried out to verify whether the GRM, which models multiple vehicles and inter-vehicle interactions, can improve the risk recognition accuracy. Lastly, the risk recognition models are trained for each of the three drivers based on the framework proposed in this paper. A five-fold cross validation is used to evaluate the performance of the proposed framework, which means that the model is trained and tested using 3640 and 910 sample scenes, respectively, and the training process is repeated five times to get the average accuracy. The two graph kernel methods are used for comparison.

*A. Experimental Setup*

We are using Naturalistic driving data collected by the Automated Vehicle Testing and Evaluation Technology Project of China Automotive Engineering Research Institute Co., Ltd. (CAERI-NDS) to validate the proposed framework. The data acquisition platform is based on the CHANGAN CS75 vehicle with various sensors installed for collecting driving scene features, as shown in Fig. 7. The platform sensors include a global positioning system (GPS), a front millimeter wave radar, left rear, and right rear). It also collects the operational data of driver through the CAN bus. The platform can continuously collect data at a high sampling rate, which is typically equal to 25 Hz, as shown in Table 1. The data include the CAN signals and the GPS position information of the platform, the relative distance, the speed and acceleration of vehicles surrounding the platform, the lane line coordinate data and the CIPV (Closet-in-Path Vehicle) tags. The data collection team consists of data collectors, support staff and three drivers with different driving experience, namely Driver A, Driver B and Driver C.

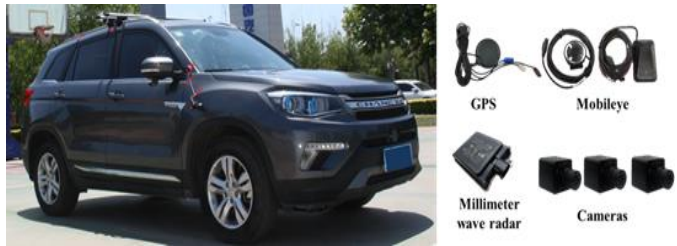

Fig. 7. Data Acquisition Platform.

Table.1 Data collection content

| Feature type | Feature name | Sensors |
|---|---|---|
| Static scenes feature | Number of lane lines<br>Position of lane lines<br>Road type | Camera |
| Dynamic scenes feature | Relative Speed of surrounding vehicle<br>Relative position of surrounding vehicle | Mobileye & Millimeter wave radar |
| Operational data | Steering wheel angle<br>Gas pedal position<br>Brake Signal<br>Steering<br>Longitudinal and lateral acceleration | CAN bus |
| Other data | Latitude and longitude of host vehicle | Differential GPS |

The total mileage of the collected real driving data is 150,000 km, which is mixed with useless scene data. Therefore, it is necessary to extract useful data to generate training and testing data sets prior to training the model. The extracted data should meet the following requirements: 1. The host vehicle stays straight ahead. 2. There are no less than two surrounding vehicles in the driver's perspective. 3. The surrounding vehicles in the driver's perspective have lane-changing behavior. In order to eliminate the noise in the collected data, the local regression using weighted linear least squares method is used to smooth the raw data as shown in Fig. 8.

**GRM-based scene features representation:** As described in section IV, first, the road structure map in the driving scenes, namely the static features, is extracted. The velocity of the host vehicle in the x and y directions is integrated to obtain the vehicle trajectory in the world coordinate system. The obtained trajectory and the detected lane coordinates transformed from the host vehicle coordinate system are combined. Subsequently, the road structure map in the world coordinate system is drawn.

Second, the relative distances between the host vehicle and the surrounding vehicles, namely the dynamic features, are extracted.

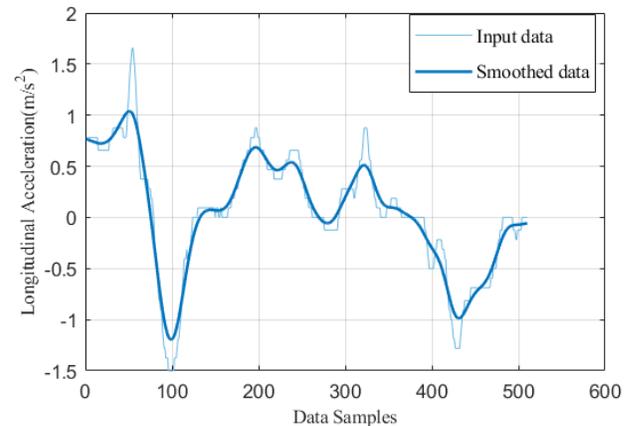

Fig. 8. Smoothing of acquisition data.

After the coordinate system conversion, the coordinates of the surrounding vehicles in each frame of the scene can be projected onto the generated road structure map. As Fig. 9 shows, the red and blue boxes represent the host vehicle and the surrounding vehicles, respectively, and all vehicles are defined as the graph nodes.

The discrete graph node labels are obtained by plotting the moving scale line, i.e., the red dashed line with the host vehicle as the origin of the coordinate system. The farthest detection distance of the platform's sensors is 100 m. Considering the vehicle length and following distance, the y-direction is divided into 10 areas with a 10 m interval. The plotted moving scale line is combined with the road structure map, and each frame of the aerial view is divided into 3x10 grids and numbered sequentially. The corresponding graph node label is obtained according to the number of the raster where the vehicle is located.

Last, the interaction between vehicles in the driving scene is modeled. The grid coordinates where each graph node is located were obtained in the previous step. In our proposed framework, the interactions between the vehicles are defined as follows:

If there are other graph nodes in the 3x3 grid range centered on the graph node A, those nodes are considered to have edge connections with the graph node A. As there is bidirectional interaction in the driving scenes, the edges $E_t$ are defined as undirected and labeled as 1. In a graph, a node is defined as a free node if it has no edges with all other nodes. This is because the vehicle represented by such a node is too far away from the other vehicles. It can be assumed that this vehicle does not interact with other vehicles and the node is removed from the graph. In summary, the GRM-based driving scene features are constructed from the collected real-world dataset. Fig. 10 demonstrates a traffic scene segment with the constructed graph data.

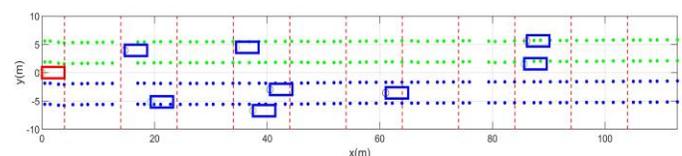

Fig. 9. Generation of aerial views for traffic scenes.

**VRM-based scene features representation:** Many existing risky scenes recognition methods use vector representation to extract feature vectors from the driving scene. These feature vectors include the trajectory of surrounding vehicles and their speeds relative to the host vehicle. However, the limitations of the vector representation form limit the ability of the extracted feature vectors to explicitly represent the interaction between the features. Meanwhile, as the number of surrounding vehicles in the driving scene is constantly changing, the extracted feature vectors are not of equal length. This is unacceptable for some widely used classifiers, such as the SVM. Therefore, existing risky scenes recognition methods usually focus on vehicles with a varying behavior around the host vehicle, such as the lane-changing vehicles, and extract their feature vectors, shown in (8) as follows:

$$\mathbf{s}_t = [\Delta x_t, \Delta y_t, \Delta v_{x,t}, \Delta v_{y,t}]^T \quad (8)$$

where $\Delta x_t$ and $\Delta y_t$ are the relative distances between the overtaking vehicle and the vehicle in the x and y directions, respectively. The relative speeds between the lane-changing vehicle and the host vehicle in the x and y directions are denoted by $\Delta v_{x,t}$ and $\Delta v_{y,t}$, respectively.

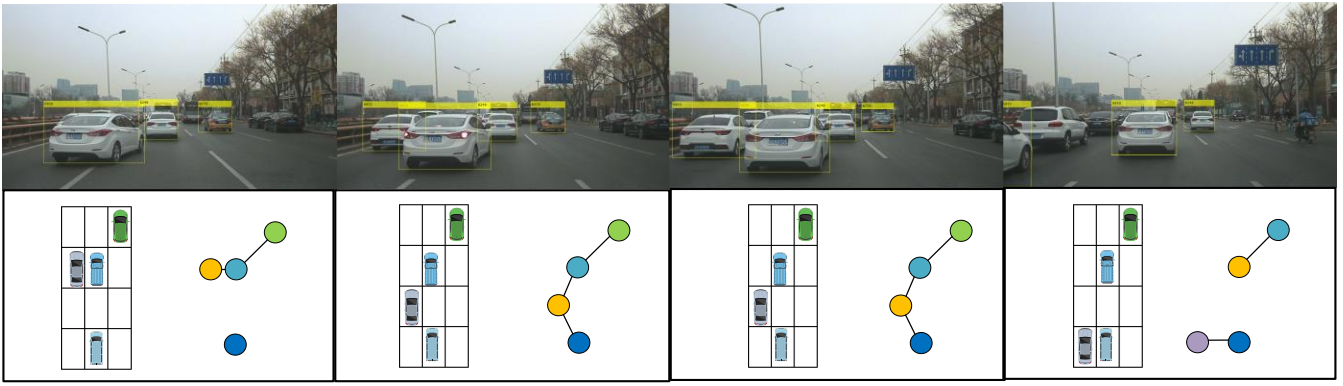

Fig. 10. A segment of traffic scene with the corresponding graph model.

*B. Personalized risk assessment label generation*

It is necessary to capture the subjective risk assessment patterns of different drivers to train a personalized risky scenes recognition model and obtain the risk level labels based on the operational data of the driver. Therefore, in this experiment, operational features having strong correlations with the risk perception of drivers are studied and compared. A clustering result that is as clearly divided as possible is obtained, and two clustering methods and a set of clustering number $K$ are tested.

First, as described in the section III, two feature vectors $\mathbf{s}_{1,t}$ (Feature One) and $\mathbf{s}_{2,t}$ (Feature Two) are constructed from the collected operation data for comparison, as shown below.

$$\mathbf{s}_{1,t} = [a_{x,t}]^T \quad (9)$$

$$\mathbf{s}_{2,t} = [a_{x,t}, a_{y,t}, \theta_t, b_t, u_t]^T \quad (10)$$

where $a_{x,t}$ and $a_{y,t}$ are the vehicle longitudinal and lateral accelerations, respectively, $\theta_t$ is the vehicle front wheel rotation angle, $b_t$ is the brake signal, and $u_t$ is the throttle opening signal. The first experiment uses operational data from the real-world dataset collected by driver A for validation. The features of $\mathbf{s}_{2,t}$ are normalized within the range -1 and 1 as follows to guarantee that all of them have the same importance:

$$\mathrm{norm}(\mathbf{s}_{2,t}) = \frac{2\left[\mathbf{s}_{2,t} - 0.5(\mathbf{s}_2^{\max} + \mathbf{s}_2^{\min})\right]}{\mathbf{s}_2^{\max} - \mathbf{s}_2^{\min}} \quad (11)$$

where $\mathbf{s}_2^{\max}$ and $\mathbf{s}_2^{\min}$ are the vectors composed of the maximum and minimum values of all features, respectively. In the following test, $\mathrm{norm}(\mathbf{s}_{2,t})$ is used to replace $\mathbf{s}_{2,t}$.

In this experiment, the KMC is first used to separately cluster $\mathbf{s}_{1,t}$ and $\mathrm{norm}(\mathbf{s}_{2,t})$, and the KPCA-KMC is also chosen to

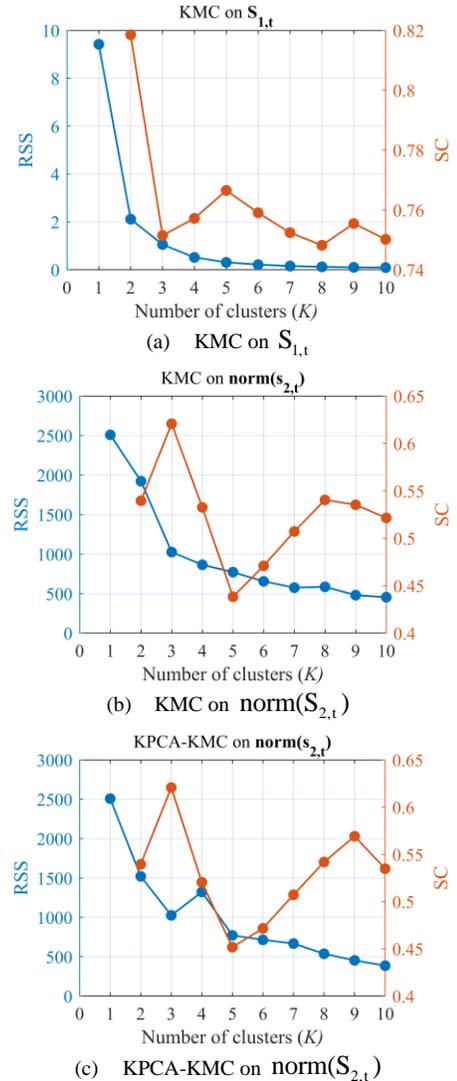

(a) KMC on $\mathbf{S}_{1,t}$

(b) KMC on $\mathrm{norm}(\mathbf{S}_{2,t})$

(c) KPCA-KMC on $\mathrm{norm}(\mathbf{S}_{2,t})$

Fig. 11. The optimal K-value selection for two features.

Table.2 The optimal k-value selection for clustering operational data of three drivers.

| Driver | | k=2 | k=3 | k=4 | k=5 | k=6 | k=7 | k=8 | k=9 | k=10 |
|---|---|---|---|---|---|---|---|---|---|---|
| **Driver A** | RSS | 7.31 | 1.06 | 0.54 | 0.21 | 0.08 | 0.06 | 0.04 | 0.02 | 0.02 |
| | SC | **0.82** | 0.75 | 0.76 | 0.77 | 0.76 | 0.75 | 0.74 | 0.76 | 0.74 |
| **Driver B** | RSS | 9.67 | 0.67 | 1.13 | 0.08 | 0.19 | 0.11 | 0.07 | 0.01 | 0.07 |
| | SC | **0.86** | 0.84 | 0.81 | 0.80 | 0.78 | 0.77 | 0.78 | 0.79 | 0.78 |
| **Driver C** | RSS | 11.17 | 2.14 | 0.61 | 0.57 | 0.15 | 0.09 | 0.08 | 0.02 | 0.06 |
| | SC | 0.76 | 0.75 | 0.72 | **0.78** | 0.75 | 0.74 | 0.72 | 0.73 | 0.73 |

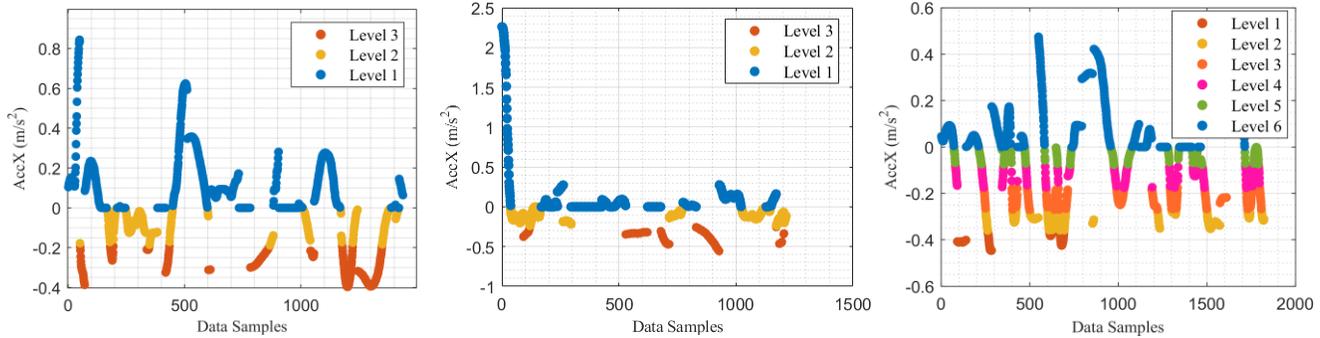

Fig. 13. The results of longitudinal acceleration clustering for three drivers.

Fig. 14. Confusion matrix for Driver A's risk scenes recognition results.

cluster $\text{norm}(\mathbf{s}_{2,t})$ considering its high dimensionality. At the same time, a set of different $K$ values is used for clustering in order to determine the number of clusters, where these values represent the perceived risk rating of driving scenes by the driver. Afterwards, the RSS and the SC are calculated for each clustering result, as shown in Fig. 11.

In order to select the optimal $K$ value, the "elbow principle" of RSS is used. The RSS value of $\mathbf{s}_{1,t}$ is shown by the blue line in Fig. 11 (a). It can be observed that the "elbow" appears when the $k$ value is 2, and the corresponding SC value is 0.82 as shown by the red line. As the $k$ value increases, there is no significant change in the RSS, and the SC is close to 1. Therefore, it can be assumed that the optimal $K$ value is 2. For $\text{norm}(\mathbf{s}_{2,t})$, as shown in Fig. 11 (b) and Fig. 11 (c), it can be observed that there is no "elbow" in the RSS values of both the KMC and KPCA-KMC, and all the SC values drop below 0.65. This behavior indicates that there is no significant differentiation between the data points of $\text{norm}(\mathbf{s}_{2,t})$.

The results of Experiment 1 illustrate that the data of $\text{norm}(\mathbf{s}_{2,t})$ do not have obvious class features and cannot be accurately clustered using the KMC and KPCA-KMC. The clustering performance of $\mathbf{s}_{1,t}$ is more suitable than that of $\text{norm}(\mathbf{s}_{2,t})$ to describe the risk assessment pattern of drivers.

One explanation for this phenomenon is that the performance of unsupervised learning is perturbed by certain dimensions in $\text{norm}(\mathbf{s}_{2,t})$ that are not relevant to the reaction of the driver in risky scenes. For further discussion, the operational data of Driver A in a risky scene are selected for presentation. The data normalization operation is first performed to eliminate the effect of different dimensional data units.

As Fig. 12 shows, when the driver encounters a risky driving scene, he/she will first release the gas pedal and then continue to press the brake pedal if the risk level of the scene does not reduce. This series of operations causes the longitudinal acceleration of the host vehicle to decrease and become negative. Thus, it demonstrates that the longitudinal acceleration can represent the gas and brake pedal operations of the driver. However, during the above process, the changes in lateral acceleration and front wheel deflection angle do not correlate with the braking operation of the driver. This shows that the driver prefers to use fast braking rather than steering to avoid potential risks. A possible reason for this choice is that In summary, it is verified that the longitudinal acceleration of the vehicle is an effective representation of the subjective perception of driver about the driving risks. Therefore, the operational data of the three drivers are clustered using the KMC to generate training labels. To determine the *K* values for each driver separately, the calculated RSS and SC values are shown in Table 2. Based on the "elbow principle" and the SC, the obtained *K* values for drivers A and B are equal to two, and the *K* value for driver C is equal to five. Driver C has the least driving experience among the three drivers; therefore, he is more likely to use the emergency brake when encountering risky scenes and is more sensitive to the change of risk level of the scenes.

After combining the "not dangerous" category with the positive longitudinal acceleration, the risk levels of drivers A and B are divided into three categories, and that of driver C is divided into six categories. Fig. 13 shows the clustering results.

*C. GRM-based Experiments Result*

In order to have an intuitive understanding of the effect of GRM on risk identification accuracy, the VRM-based and GRM-based risk recognition models for driver A are trained separately, and their recognition accuracies are compared. Two graph kernels, namely SPGK and NHGK are used to train the classifiers for the latter model. Thus, there are a total of three trained classifiers, which are denoted as: 1) $\Phi_{SPGK}$ where the training data are the graph feature set $S_G$ combined with the graph kernel method SPGK, 2) $\Phi_{NHGK}$ where $S_G$ is combined with the graph kernel method NHGK, 3) $\Phi_{LC}$ where the training data are the state feature set $S_{LC}$ of the lane changing vehicle, directly combined with the linear SVM. As different data units in $S_{LC}$ will lead to different data weights during the training process, each feature of $S_{LC}$ is normalized separately. Three SVM classifiers are trained and evaluated using five-fold cross-validation, and the average of five experimental results is taken as the final result to eliminate the effect of random error. The confusion matrices of the recognition results of $\Phi_{SPGK}$, $\Phi_{NHGK}$ and $\Phi_{LC}$ for driver A are plotted in Fig. 14.

The results show that the overall accuracy rates of $\Phi_{SPGK}$, $\Phi_{NHGK}$ and $\Phi_{LC}$ are 89.9%, 89.5% and 68.1%, respectively. These results indicate that the GRM used in this framework is generally better than the VRM in the driving scene feature extraction process. As the proposed framework aims to enhance driving safety, special attention should be paid to the accuracy of the "risk levels", i.e., "level1" and "level2" in this example. It can be noted from the target class bar, shown in gray color at the bottom of the confusion matrix, that the accuracy rates of "level1" (class 1) and "level2" (class 2) are 35.3% and 30.2%, respectively for $\Phi_{LC}$, 80.8% and 80.1%, respectively for $\Phi_{SPGK}$, and 73.9% and 83.6%, respectively for $\Phi_{NHGK}$.

It is obvious that the accuracy of $\Phi_{LC}$ is considerably lower than that of $\Phi_{SPGK}$ and $\Phi_{NHGK}$. A possible reason is the difficulty in fully representing the hidden interactions between vehicles in real driving scenes by only extracting vector information of lane changing vehicles. The risk level of the scenes is influenced by a combination of various factors such as the changing position of multiple surrounding vehicles and the right-of-way game between vehicles. Therefore, this framework uses the GRM that can effectively characterize the number of changing vehicles and the hidden interaction between the vehicles, which is verified by the accuracy of $\Phi_{SPGK}$ and $\Phi_{NHGK}$.

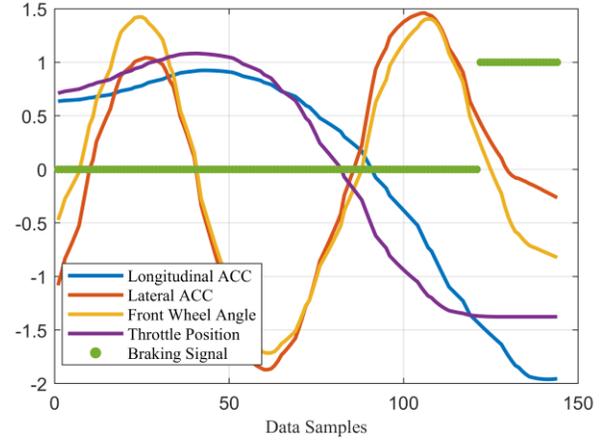

Fig. 12. A segment of driver operation data.

However, the present framework also has its shortcomings. It can be observed that $\Phi_{SPGK}$ and $\Phi_{NHGK}$ tend to recognize level1 as level2 with error rates of 4.3% and 5.6%, respectively, which may cause safety issues. However, as shown above, the classification accuracy improves further as the number of samples increases, which can reduce the false recognition results to some extent. However, the recognition errors still exist in our framework, which is due to our limited understanding of the interaction behavior between vehicles.

Based on this observation, further research on inter-vehicle interactions will be added to our system by carefully defining the graph edges as a basis for future risk assessment.

*D. Driver-Specific Risky Scene Recognition Result*

Finally, the driver-specific risky scene recognition performance of the proposed framework is evaluated. This evaluation is carried out for each driver via GRM-based experiments with two graph kernel methods and VRM-based experiments. All three classifiers use the risk level label set $\mathbf{L}_{acc}$ as the training label, obtained from the lateral acceleration clustering results of each driver. Fig. 15 shows the results of the risky scenes recognition experiments.

The classifiers $\Phi_{SPGK}$ and $\Phi_{NHGK}$ for all three drivers have accuracy rates of 80% or higher. Driver B has the highest accuracy rates of $\Phi_{SPGK}$ and $\Phi_{NHGK}$, reaching 95.8% and 93.8%, respectively. However, the accuracy of driver C is lower compared to that of drivers A and B. This may be explained by the fact that driver C has the highest number of evaluation levels equal to six, and the collected data are not enough for the model to learn the characteristics of each risk category. Despite this, the accuracy values of $\Phi_{SPGK}$ and $\Phi_{NHGK}$ for driver C reach 81.8% and 80.1%, respectively. This indicates that the proposed framework in this paper can accurately recognize risky scenes for different drivers.

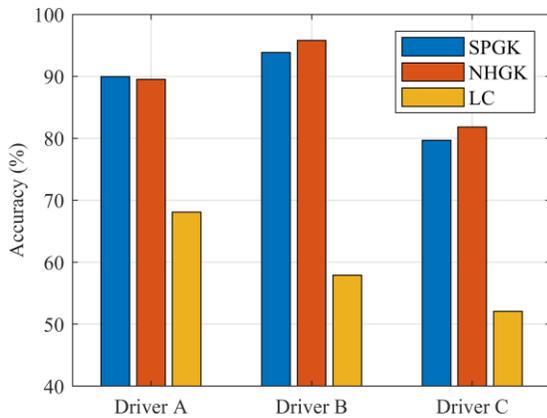

Fig. 15. The recognition accuracy of risky scenes.

VI. CONCLUSION

In this paper, a driver-specific risk recognition framework for autonomous vehicles was proposed. The framework could extract inter-vehicle interactions from urban driving scenarios. Experiments were conducted using real-world urban driving dataset collected by a total of three drivers, referred to as Drivers A, B and C to evaluate the performance of the proposed framework. Experimental results proved that the proposed framework could output the risk levels of driving scenes according to the subjective risk perceptions of the drivers, which provided a new idea for the development of personalized assisted driving systems. Besides, the proposed driver-specific risk label generation method was able to avoid both problems of high labor cost of existing subjective labeling methods and poor adaptability of objective risk evaluation indicators. Moreover, compared with the existing VRM-based method, the proposed GRM-based method could better represent the dynamic and complex of driving scenes in urban environments. All three drivers achieved a high level of risky scenes recognition accuracy. However, compared to the other two drivers, the recognition accuracy of Driver C could still be improved. This may be because this driver was more sensitive to risks and had a denser grading of risky levels, which led to an uneven distribution of training data. Therefore, further exploration of the recognition performance of the proposed framework to find theoretically effective critical training data amount will be carried out in future.

REFERENCES

[1] S. Singh, "Critical reasons for crashes investigated in the national motor vehicle crash causation survey," 2015.
[2] E. Petridou and M. Moustaki, "Human factors in the causation of road traffic crashes," *European journal of epidemiology,* vol. 16, no. 9, pp. 819-826, 2000.
[3] D. N. Lee, "A theory of visual control of braking based on information about time-to-collision," *Perception,* vol. 5, no. 4, pp. 437-459, 1976.
[4] R. G. FULLER, "Determinants of time headway adopted by truck drivers," *Ergonomics,* vol. 24, no. 6, pp. 463-474, 1981.
[5] O. Khatib, "Real-time obstacle avoidance for manipulators and mobile robots," in *Autonomous robot vehicles*: Springer, 1986, pp. 396-404.
[6] J. Li, H. Ma, Z. Zhang, and M. Tomizuka, "Social-WaGDAT: Interaction-aware Trajectory Prediction via Wasserstein Graph Double-Attention Network," *arXiv preprint arXiv:.06241,* 2020.
[7] X. Li *et al.*, "Vehicle Trajectory Prediction Using Generative Adversarial Network With Temporal Logic Syntax Tree Features," vol. 6, no. 2, pp. 3459-3466, 2021.
[8] J. Hu, X. Zhang, and S. Maybank, "Abnormal Driving Detection With Normalized Driving Behavior Data: A Deep Learning Approach," *IEEE Transactions on Vehicular Technology,* vol. 69, no. 7, pp. 6943-6951, 2020.
[9] Y. Xing *et al.*, "Driver Lane Change Intention Inference for Intelligent Vehicles: Framework, Survey, and Challenges," *IEEE Transactions on Vehicular Technology,* pp. 1-1, 2019.
[10] H. Huang *et al.*, "A probabilistic risk assessment framework considering lane-changing behavior interaction," *Science China Information Sciences,* vol. 63, no. 9, p. 190203, 2020/08/17 2020.
[11] A. Aksjonov, P. Nedoma, V. Vodovozov, E. Petlenkov, and M. Herrmann, "Detection and Evaluation of Driver Distraction Using Machine Learning and Fuzzy Logic," *IEEE Transactions on Intelligent Transportation Systems,* vol. 20, no. 6, pp. 2048-2059, 2019.
[12] E. Yurtsever *et al.*, "Integrating driving behavior and traffic context through signal symbolization for data reduction and risky lane change detection," *IEEE Transactions on Intelligent Vehicles,* vol. 3, no. 3, pp. 242-253, 2018.
[13] Q. Shangguan, T. Fu, J. Wang, and T. Luo, "An integrated methodology for real-time driving risk status prediction using naturalistic driving data," *Accident Analysis & Prevention,* vol. 156, p. 106122, 2021.
[14] M. Li, S. Chen, X. Chen, Y. Zhang, Y. Wang, and Q. Tian, "Symbiotic graph neural networks for 3d skeleton-based human action recognition and motion prediction," *IEEE Transactions on Pattern Analysis Machine Intelligence,* 2021.
[15] J. Zürn, J. Vertens, and W. Burgard, "Lane Graph Estimation for Scene Understanding in Urban Driving," *arXiv preprint arXiv:.00195,* 2021.
[16] Y. Gao, Y.-F. Li, Y. Lin, H. Gao, and L. Khan, "Deep learning on knowledge graph for recommender system: A survey," *arXiv preprint arXiv:2004.00387,* 2020.
[17] Y. Qiao, X. Luo, C. Li, H. Tian, and J. Ma, "Heterogeneous graph-based joint representation learning for users and POIs in location-based social network," *Information Processing & Management,* vol. 57, no. 2, p. 102151, 2020.
[18] S. Mylavarapu, M. Sandhu, P. Vijayan, K. M. Krishna, B. Ravindran, and A. Namboodiri, "Understanding dynamic scenes using graph convolution networks," in *2020 IEEE/RSJ International Conference on Intelligent Robots and Systems (IROS)*, 2020, pp. 8279-8286: IEEE.
[19] J. Li, F. Yang, M. Tomizuka, and C. Choi, "Evolvegraph: Heterogeneous multi-agent multi-modal trajectory prediction with evolving interaction graphs," *ArXiv, abs/2003.13924,* vol. 2, 2020.
[20] B. Zhu, J. Han, J. Zhao, and H. Wang, "Combined hierarchical learning framework for personalized automatic lane-changing," *IEEE Transactions on Intelligent Transportation Systems,* 2020.
[21] Y. Ren, S. Elliott, Y. Wang, Y. Yang, and W. Zhang, "How Shall I Drive? Interaction Modeling and Motion Planning towards Empathetic and Socially-Graceful Driving," in *2019 International Conference on Robotics and Automation (ICRA)*, 2019, pp. 4325-4331.
[22] B. Zhu, S. Yan, J. Zhao, and W. Deng, "Personalized lane-change assistance system with driver behavior identification," *IEEE Transactions on Vehicular Technology,* vol. 67, no. 11, pp. 10293-10306, 2018.
[23] P. Ping, Y. Sheng, W. Qin, C. Miyajima, and K. Takeda, "Modeling driver risk perception on city roads using deep learning," *IEEE Access,* vol. 6, pp. 68850-68866, 2018.
[24] A. Siren and M. R. Kjær, "How is the older road users' perception of risk constructed?," *Transportation research part F: traffic psychology and behaviour,* vol. 14, no. 3, pp. 222-228, 2011.
[25] D. Crundall *et al.*, "Some hazards are more attractive than others: Drivers of varying experience respond differently to different types of hazard," *Accident Analysis & Prevention,* vol. 45, pp. 600-609, 2012.
[26] J. Wang, H. Huang, Y. Li, H. Zhou, J. Liu, and Q. Xu, "Driving risk assessment based on naturalistic driving study and driver attitude

[27] questionnaire analysis," *Accident Analysis & Prevention,* vol. 145, p. 105680, 2020.
[27] J. Wang, Y. Zheng, X. Li, C. Yu, K. Kodaka, and K. Li, "Driving risk assessment using near-crash database through data mining of tree-based model," *Accident Analysis & Prevention,* vol. 84, pp. 54-64, 2015.
[28] F. Guo, S. G. Klauer, J. M. Hankey, and T. A. Dingus, "Near crashes as crash surrogate for naturalistic driving studies," *Transportation Research Record,* vol. 2147, no. 1, pp. 66-74, 2010.
[29] M. Cai *et al.*, "The association between crashes and safety-critical events: Synthesized evidence from crash reports and naturalistic driving data among commercial truck drivers," *Transportation research part C: emerging technologies,* vol. 126, p. 103016, 2021.
[30] S. S. Pantangi, G. Fountas, P. C. Anastasopoulos, J. Pierowicz, K. Majka, and A. Blatt, "Do High Visibility Enforcement programs affect aggressive driving behavior? An empirical analysis using Naturalistic Driving Study data," *Accident Analysis & Prevention,* vol. 138, p. 105361, 2020.
[31] G. Xie, H. Qin, M. Hu, D. Ni, and J. Wang, "Modeling discretionary cut-in risks using naturalistic driving data," *Transportation research part F: traffic psychology and behaviour,* vol. 65, pp. 685-698, 2019.
[32] K. P. Sinaga and M.-S. Yang, "Unsupervised K-means clustering algorithm," *IEEE Access,* vol. 8, pp. 80716-80727, 2020.
[33] F. Anowar, S. Sadaoui, and B. Selim, "Conceptual and empirical comparison of dimensionality reduction algorithms (PCA, KPCA, LDA, MDS, SVD, LLE, ISOMAP, LE, ICA, t-SNE)," *Computer Science Review,* vol. 40, p. 100378, 2021.
[34] S. Ulbrich, T. Menzel, A. Reschka, F. Schuldt, and M. Maurer, "Defining and substantiating the terms scene, situation, and scenario for automated driving," in *2015 IEEE 18th International Conference on Intelligent Transportation Systems*, 2015, pp. 982-988: IEEE.
[35] K. M. Borgwardt and H.-P. Kriegel, "Shortest-path kernels on graphs," in *Fifth IEEE international conference on data mining (ICDM'05)*, 2005, p. 8 pp.: IEEE.
[36] S. Hido and H. Kashima, "A linear-time graph kernel," in *2009 Ninth IEEE International Conference on Data Mining*, 2009, pp. 179-188: IEEE.
[37] J. C. Gower, "A general coefficient of similarity and some of its properties," *Biometrics,* pp. 857-871, 1971.
[38] C.-C. Chang and C.-J. Lin, "LIBSVM: a library for support vector machines," *ACM transactions on intelligent systems and technology (TIST),* vol. 2, no. 3, pp. 1-27, 2011.